\documentclass{sig-alternate-mod}
\usepackage{flushend}
\usepackage{cite}
\usepackage{algorithm}
\usepackage[noend]{algorithmic}
\usepackage{colortbl}
\usepackage{subfig}
\definecolor{res2}{RGB}{255,236,139}
\definecolor{res1}{RGB}{255,185,15}
\definecolor{res3}{RGB}{102,205,0}
\definecolor{res4}{RGB}{152,251,152}
\definecolor{olive}{RGB}{85,107,47}
\definecolor{res}{RGB}{153,149,140}
\newcommand{\ind}{\parindent=-0.15cm}
\graphicspath{{pic/}}
\usepackage{graphicx}


\begin{document}
%

\title{Adaptive Parameter Selection in Evolutionary Algorithms \\by Reinforcement Learning with Dynamic Discretization \\of Parameter Range}
\numberofauthors{3}
\author{
\alignauthor Arkady Rost\\
\affaddr{ITMO University}\\
\affaddr{49 Kronverkskiy ave.}\\
\affaddr{Saint-Petersburg, Russia}\\
\email{arkrost@gmail.com}
\alignauthor Irina Petrova\\
\affaddr{ITMO University}\\
\affaddr{49 Kronverkskiy ave.}\\
\affaddr{Saint-Petersburg, Russia}\\
\email{petrova@rain.ifmo.ru}
\alignauthor Arina Buzdalova\\
\affaddr{ITMO University}\\
\affaddr{49 Kronverkskiy ave.}\\
\affaddr{Saint-Petersburg, Russia}\\
\email{abuzdalova@gmail.com}       
}


\maketitle

\begin{abstract}

Online parameter controllers for evolutionary algorithms adjust values of parameters during the run of an evolutionary algorithm. 
Recently a new efficient parameter controller based on reinforcement learning was proposed by Karafotias et al. 
In this method ranges of parameters are discretized into several intervals before the run. However, performing adaptive discretization during the run 
may increase efficiency of an evolutionary algorithm. Aleti et al. proposed another efficient controller with adaptive discretization. 

In the present paper we propose a parameter controller based on 
reinforcement learning with adaptive discretization. The proposed controller is compared with the existing parameter 
adjusting methods on several test problems using different configurations of an evolutionary algorithm. For the test 
problems, we consider four continuous functions, namely the sphere function, the Rosenbrock function, the Levi function 
and the Rastrigin function. 
Results show that the new controller outperforms the other controllers on most of the considered test problems.

\end{abstract}



%
%

\section{Introduction}

Let us denote efficiency of an evolutionary algorithm (EA) as the number of fitness function evaluations needed to find the optimal solution. Efficiency of EA 
is strongly correlated with its parameters. Common examples of such parameters are mutation and crossover probabilities. Optimal values of the parameters not only 
depend on the type of EA but also on the characteristics of the problem to be solved. Values of the parameters can be set before a run. However, optimal 
parameter values can change during a run, so an approach for adaptive parameter adjustment is required.  


We consider parameters with continuous values. When adjusting such parameters, parameter ranges are usually discretized 
into some intervals. We can discretize parameter ranges a priori and keep the resulting segmentation during a run.  
However, it was shown that adaptive discretization during optimization process may improve the performance of the 
algorithm~\cite{arpc, earpc}.
A possible explanation of this fact is as follows. The dynamic discretization allows to split the intervals into smaller subintervals. If the size of an interval is small, it is more likely to choose a good parameter value. 
Aleti et al. proposed entropy-based adaptive range parameter controller (EARPC)~\cite{earpc}. This method uses adaptive discretization. 

Recently Karafotias et al. proposed another efficient parameter controller based on reinforcement learning~\cite{karafotias-gecco, karafotias-overview}. 
However, this method was not compared to EARPC. In the method proposed by Karafotias et al. a priori discretization is used. A new method which combines usage of 
reinforcement learning and dynamic discretization is proposed in this work. 

The rest of the paper is organized as follows. First, the basic ideas used in EARPC 
and the method proposed by Karafotias are described. It is necessary to describe these ideas in order to explain how the proposed controller works. Second, two different versions of a new controller are proposed. Then the experiments are described. Finally, the new parameter controllers are compared with the other considered controllers. 



%

\section{Related work}

Let us give the formal description of the adaptive parameter control problem. We have a set of $n$ parameters $v_1, ..., v_n$. The goal of the parameter controller is to
select parameter values which maximize efficiency of EA. The first parameter controller to be considered is EARPC.
\subsection{Entropy-based adaptive range parameter controller}

The EARPC method~\cite{earpc} adjusts parameters separately. 
Let us denote a set of the selected parameter values as $v = (v_1, \ldots, v_n)$. The efficiency of EA with parameters set 
to $v$ is denoted as $q(v)$. During a run of the algorithm we save pairs of $v$ and $q(v)$. 

To select a new set of parameter values, we split the 
saved values of $v$ into two clusters. For example, it can be done by k-means algorithm. Next, the range of each 
parameter $v_i$ is split into two 
intervals. To select a split point, all saved values of $v_i$ are sorted in ascending order. Candidates to be a split point are mid-points between two 
consecuent values of $v_i$ in the sorted list. For each candidate $s_k$ to be a split point, the set of the saved values 
of $v_i$ is split into two subsets 
$p_1$ and $p_2$. The subset $p_1$ contains values which are less than or equal to $s_k$. The subset $p_2$ contains values greater than $s_k$. Denote the number 
of the saved values of $v_i$ which are contained in the cluster $c_i$ and the subset $p_j$ as $c_i(p_j)$. The entropy $H$ is calculated according to 
Eq.~\ref{entropy} for each candidate $s_k$.

\begin{align}
\label{entropy}
e_{p_1} & = -\frac{|c_1(p_1)|}{|p_1|}\ln\left(\frac{|c_1(p_1)|}{|p_1|}\right) -\frac{|c_2(p_1)|}{|p_1|}\ln\left(\frac{|c_2(p_1)|}{|p_1|}\right), \nonumber \\
e_{p_2} & = -\frac{|c_1(p_2)|}{|p_2|}\ln\left(\frac{|c_1(p_2)|}{|p_2|}\right) -\frac{|c_2(p_2)|}{|p_2|}\ln\left(\frac{|c_2(p_2)|}{|p_2|}\right), \\
H & = \frac{|p_1|}{|c_1|}e_{p_1} + \frac{|p_2|}{|c_2|}e_{p_2} \nonumber 
\end{align}
The split point $s$ with the minimal value of entropy is selected. Two intervals $[min, s]$ and $(s, max]$ are 
obtained, where $min$ and $max$ are 
the minimal and the maximal possible values of $v_i$ correspondingly. Values from the set $p_1$ correspond to the 
first interval, values from the set $p_2$ 
correspond to the second interval. 

To decide from which interval we should choose a new value for the parameter $v_i$, we calculate the average quality 
of the parameter values in each interval. 
Let $Q_1$ and $Q_2$ denote the average quality of the values in the first and the second intervals correspondingly. 
$Q_j$ is calculated according 
to Eq.~\ref{avquality}.
\begin{align}
\label{avquality}
Q_j & = \frac{1}{|p_j|}\sum\limits_{v \in p_j}q(v)
\end{align}
Then we randomly select interval, the probability of selection of interval $j$ is proportional to $Q_j$. A new 
value of $v_i$ is 
randomly chosen from the selected interval. The pseudocode of EARPC is presented in Algorithm~\ref{earpclisting}. We 
wrote this pseudocode based on the description of EARPC given 
in~\cite{earpc}.  

\begin{algorithm}[h]
    \caption{\textit{EARPC} algorithm proposed by Aleti et al.}
    \label{earpclisting}
    \begin{algorithmic}[1]
        \STATE {Earlier selected and saved values of $v$ are split into two clusters $c_1$ and $c_2$}
	\FOR {each parameter $v_i$}
	    \STATE {Sort saved values of $v_i$ in ascending order}
	    \STATE {$H_{best} \gets \infty$}
	    \FOR {split point $s_k = \frac{v_{ij} + v_{i(j+1)}}{2}$}
	      \STATE {Split saved values of $v_i$ into two sets $p_1$ and $p_2$ according to $s_k$}
	      \STATE {Calculate entropy \textit{H} according to Eq.~\ref{entropy}} 
	      \IF {$H_{best} < H$}
		\STATE {$H_{best} \gets H$}
		\STATE {$s \gets s_k$}
	      \ENDIF
	    \ENDFOR
	    \STATE {Split saved values of $v_i$ into two sets $p_1$ and $p_2$ according to $s$}
	    \STATE {$Q_1 = \frac{1}{|p_1|}\sum\limits_{v \in p_1}{q(v)}$, $Q_2 = \frac{1}{|p_2|}\sum\limits_{v \in p_2}{q(v)}$}
	    \STATE {Randomly select interval, the probability of selection of interval $[min, s]$ is proportional to 
$Q_1$,  the probability of selection of 
interval $(s, max]$ is proportional to $Q_2$}
	    \STATE {Randomly select value of $v_i$ from the selected interval}
	\ENDFOR
    \end{algorithmic}
\end{algorithm}

\subsection{Parameter selection by reinforcement learning} \label{karafotiassection}
Let us describe the general scheme of using reinforcement learning for parameter control in EA~\cite{ea-rl, op, param-control-rl, chen-2005}.
In reinforcement learning (RL)~\cite{sutton,gosavi}, the agent applies an action to the environment, then the environment returns some representation of its 
state and a numerical reward to the agent, and the process repeats. The goal of RL is to maximize the total reward~\cite{sutton}.


While adjusting parameter of EA by RL, EA is treated as an environment. An action is selection of parameter values. EA generates new 
population using the selected parameter values. The obtained reward is based on the difference of the maximal fitness in two sequential iterations. 

\begin{figure}
    \centering
    \includegraphics[width=0.30\textwidth]{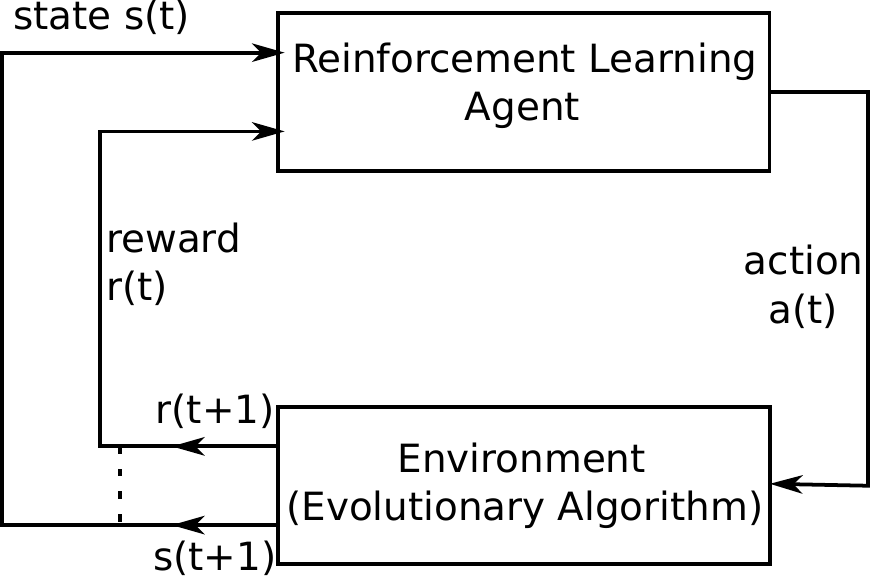}
    \caption{Reinforcement learning scheme}
        \label{rl-scheme}
\end{figure}

Let us describe how the agent selects parameter values. The parameter range is discretized into several intervals. Each interval corresponds to agent's action. 
To apply an action, the agent selects an interval and sets a random value from this interval as the parameter value. The method is illustrated in 
Fig.~\ref{rl-scheme}, where $t$ is the number of the current iteration.   

In the method proposed by Karafotias et al.~\cite{karafotias-gecco, karafotias-overview}, a modification of the $\varepsilon$-greedy Q-learning algorithm is 
used.
Let $k$ denote the number of parameters being adjusted. The range of parameter $v_i$ is 
discretized a priori into $m_i$ intervals. An action of the agent consists of selection of intervals for each parameter and random selection of parameters 
values from the selected intervals. Thus the number of possible actions of the agent is $\prod\limits_{i = 1}^k{m_i}$.   
The obtained reward is calculated according to Eq.~\ref{reward}, where $f_t$ is the best fitness function value obtained on the iteration $t$ and $c$ is a 
constant.
\begin{equation}
\label{reward}
r = c \cdot \left(\frac{f_{t+1}}{f_t} - 1\right)
\end{equation}
Note that reward is always positive unless EA worsens the best obtained solution. To reduce the learning rate in the 
case of zero reward, the learning rate $\alpha$ is changed to $\alpha_0$ according to Eq.~\ref{alphacoef}. Note that $\alpha_0 \ll \alpha$.
\begin{equation}
\label{alphacoef}
\alpha(r) = \begin{cases}\alpha, &\text{if } r > 0 \\ \alpha_0 \ll \alpha, &\text{in other cases} \end{cases}
\end{equation}

To define the state of the environment, the following observables derived from the state of the EA are used~\cite{karafotias-2012}:
\begin{itemize}
    \item genotypic diversity;
    \item phenotypic diversity (when different from genotypic);
    \item fitness standard deviation;
    \item stagnation counter (the number of iterations without fitness improvement);
    \item fitness improvement.
\end{itemize}


The dynamic state space segmentation method is used~\cite{utree}. In this method, states are represented as a binary decision tree. Each internal node contains 
a condition on an observable. Leaf nodes represent environment states. For each state $s$ array of $Q(s, a)$ is stored, where $Q(s, a)$ is the expected reward 
for each action in the state $s$. The expected reward in this state is denoted as $V(s) = \max \limits_a Q(s,a)$.   

Initially, the tree consists of one leaf which corresponds to a single state $s$ and $V(s) = 0$. Each iteration of the algorithm consists 
of two phases: the data gathering phase and the processing phase. In the data gathering phase we obtain values of observables $I = \{o_1 \ldots o_m\}$ from EA, 
where $m$ is the number of observables. Then we go down the tree using $I$ and get the state $s$ of the environment. The agent selects an action $a$ using 
$\varepsilon$-greedy strategy, obtains reward $r$ and new values $I'$ of observables. Then the agent refreshes $Q(s, a)$ and stores the resulting tuple $(I, a, 
I', r)$.

In the processing phase, we try to split the state $s$ into two new states, which means that the leaf 
corresponding to the state $s$ becomes an internal node and two new leaves are added as its children. To convert the leaf into a decision node, we have to 
choose an observable and a splitting value. For each saved tuple $(I, a, I', r)$ in the leaf we calculate an estimated reward obtained after applying the 
action $a$ to EA with the values $I$ of observables. We denote this reward as $q(I, a)$. The value of $q(I, a)$ is calculated as $q(I, a) =  r + \gamma V(s')$, 
where $s'$ corresponds to the values $I'$ of observables, and $\gamma$ is a constant called \textit{discount factor}. For each observable $o$ the tuples saved 
in the leaf are sorted in ascending order according to the value of $o$ taken from $I$. Candidates to be chosen as a split point are mid-points between two 
consecuent values of $o$ in the sorted list. The saved values $I$ taken from tuples stored in the leaf are divided into 
two subsets according to 
a candidate split point. We form two samples by dividing the calculated $q(I, a)$ according to obtained subsets of $I$. A Kolmogorov-Smirnov criterion is 
run on these two samples and the obtained p-value is saved. After all split point candidates for all observables are checked, the smallest obtained p-value is 
selected. If it is smaller than 0.05, then the node is split at the corresponding observable and the corresponding 
split point. 

The authors of the article where dynamic state space segmentation method was proposed~\cite{utree} suggest that $Q$ values should be recalculated after each 
split of a state. 
However, it is not obvious how to do it. So in the method proposed by Karafotias et al. $Q$ and $V$ values of two new 
nodes are set to the 
values of the parent node. The tuples saved in the parent node are split according to the chosen split point and the 
resulting parts are given to the 
corresponding 
children nodes. The pseudocode of the algorithm proposed by Karafotias et al. is presented in Algorithm~\ref{karafotiaslisting}. We wrote this pseudocode based 
on the description of this algorithm given in~\cite{karafotias-gecco}.

\begin{algorithm}[h]
    \caption{Algorithm proposed by Karafotias et al.}
    \label{karafotiaslisting}
    \textbf{Data gathering phase}
    \begin{algorithmic}[1]
        \STATE {Initialize binary search tree: state $s$, $V(s) = 0$, $Q(s, a) \leftarrow 0$ for each action $a$}
        \STATE {Obtain $I$ --- values of observables from EA}
        \STATE {Go down the tree using $I$ and find environment state $s$}
        \STATE {Select action $a$ using $\varepsilon$-greedy strategy}
        \STATE {Apply action $a$ to environment, obtain reward $r$}
        \STATE {Obtain $I'$ --- values of observables from EA}
        \STATE {Go down the tree using $I'$ and find environment state $s'$}
        \STATE {Store tuple $(I, a, I', r)$ in state $s$}
        \STATE {$Q(s, a) \gets Q(s,a)+\alpha(r+\gamma\max\limits_{a'}Q(s',a')-Q(s,a))$}
        \STATE {$V(s) = \max\limits_a{Q(s, a)}$}
    \end{algorithmic}
    \textbf{Processing phase}
    \begin{algorithmic}[1]
            \FOR {each tuple $(I, a, I', r)$ in state $s$}
                \STATE {Calculate $q(I, a) = r + \gamma V(s')$}
            \ENDFOR
	    \STATE {$best  \gets +\infty$}
	    \FOR {observable $o$}
		\STATE {Sort tuples stored in state $s$ according to value of $o$ from $I$.} 
		\STATE {$tuples\_count \gets$ number of tuples}
		\FOR {$j \gets 1$ to $tuples\_count$}
			\STATE {$I_j \gets$ j-th tuple in the sorted list}
		        \STATE {$o_j \gets$ value of observable $o$ in $I_j$}
			\STATE {$candidate \gets \frac{o_j + o_{j + 1}}{2}$}
			\STATE {$x \gets \{ q(I_j, a) | o_j \leq candidate\}$}
			\STATE {$y \gets \{ q(I_j, a) | o_j > candidate\}$}
			\STATE {Calculate $p$-value for $x$ and $y$ using Kolmogorov-Smirnov criterion}
			\IF {$p$-value$ < best$}
				\STATE $best \gets p$-value
				\STATE $best\_observable \gets o$
				\STATE $best\_split \gets candidate$
			\ENDIF
		\ENDFOR	
	    \ENDFOR
	    \IF {$best < 0.05$}
		    \STATE {Create two new states $s_1$ and $s_2$}
		    \STATE {Split tuples stored in \textit{s} to $s_1$ and $s_2$ corresponding to $best\_observable$ and $best\_split$}
		    \STATE {Copy $Q(s, a)$ into $Q(s_1, a)$ and $Q(s_2, a)$}
		    \STATE {Replace node corresponding to state $s$ with internal node with two new leafs, corresponding to states $s_1$ and $s_2$}
	    \ENDIF
%
%
    \end{algorithmic}
\end{algorithm}

\section{Proposed methods}
We propose two new controllers. The first of them combines the EARPC algorithm and the approach proposed by Karafotias et al. The second one is based on 
reinforcement learning and splitting of parameter ranges using Kolmogorov-Smirnov criterion.
\subsection{Method which combines Karafotias et al. and EARPC}
In the method proposed by Karafotias et al., dynamic state space segmentation is used. However, ranges of parameters being adjusted are discretized a priori. 
We propose to improve the method proposed by Karafotias et al. by using EARPC method for dynamic discretization of ranges of parameters. When we change 
discretization of the parameter range, we change the set of agent actions. So we have to change the process of selection of the parameter values in the method 
proposed by Karafotias et al. 

The values of parameters are selected as in the EARPC method. Therefore, on each iteration of the algorithm we obtain values $I$ of observables of EA and 
go down the binary decision tree of states (see Section~\ref{karafotiassection}) to find a state $s$ corresponding to $I$. Then we select 
values $v$ of parameters, get new values of observables $I'$ and save the tuple $(I, v, I', r)$ in the state $s$, where $v$ are selected values of parameters 
and $r$ is calculated according to Eq.~\ref{reward}. The values $v$ of parameters are selected by EARPC method using the tuples saved earlier in the state $s$. 
To apply the EARPC algorithm, we have to calculate $q(v)$ which is the efficiency measure of EA with the parameters set to $v$. We use $r$ from the tuple $(I, 
v, I', r)$ as the efficiency measure $q(v)$.  


To split the states, we have to calculate $q(I, a) = r + \gamma V(s')$, where $s'$ corresponds to $I'$. In the proposed method actions of the agent are 
changing during the run. So we cannot calculate $V(s')$ as $\max \limits_a Q(s',a)$. In this case we use the expected value of the reward obtained in the 
state $s'$ as $V(s')$. The EARPC algorithm splits the parameter range into two intervals. One of these intervals is 
selected with probability 
proportional to average reward on this interval. So $V(s')$ is calculated as $\sum\limits_{i = 1}^2{\frac{Q_i^2}{Q_1 + 
Q_2}}$, where $Q_1$ and $Q_2$ are 
average rewards on the two intervals.

To the best of our knowledge, there is no specific method of recalculating $Q$-values after splitting a state. In the 
method proposed by Karafotias et al., $Q$ and $V$ values of two new nodes are set to the values of the parent node. In this method we do not store $Q$-values 
in leafs. So we do not need to recalculate $Q$ and $V$ for two new states after splitting.   

\subsection{Method with adaptive selection of action set}
Preliminary experiments showed that there was no significant improvement of the EA efficiency when high number of states was used. Thus the second proposed 
method described in this section does not use the binary decision tree of states, although it is used in the first proposed method and the method proposed by 
Karafotias et al. In the second proposed method we have a single state of the environment. In the method proposed by Karafotias et al., values of all 
parameters to be adjusted are set simultaneously by the $Q$-learning agent. In the second proposed method the value of each parameter is set independently of 
the other parameters. So we have a separate $Q$-learning agent for each parameter.

Initially, for each parameter $v_i$ we have only one action of the $\text{agent}_i$ which corresponds to the selection of parameter value from the range $[min, 
max]$, where $min$ and $max$ are the minimal and the maximal possible values of the parameter $v_i$. Each $\text{agent}_i$ applies this action and sets the 
value of $v_i$. Then we use the selected values $v = (v_1 \ldots v_n)$ in EA and calculate the reward $r$ according to Eq.~\ref{reward}. We store the tuple of 
the selected values $v$ and the obtained reward $r$. These tuples are used for splitting the parameter range and, as a consecuence, they are used for 
changing sets of actions of the agents. The agents apply the single possible action until enough tuples for splitting of the range are stored. 

After enough tuples are stored, we search a split point for the range of each parameter using 
Kolmogorov-Smirnov criterion. The criterion is used in 
the same way as it was used by Karafotias et al. when splitting states. If the split point is not found, we do not split the range. If the split point is 
found, we obtain two intervals $L$ and $R$. Then we try to split $L$ and $R$ in the same way. We repeat this process 
$i$ times. So the maximum number of the 
intervals is $2^i$. 

An agent selects an action using $\varepsilon$-greedy strategy until $Q$-values become almost equal for all actions. In this case the expected rewards for all 
actions are almost equal, therefore the agent can not select which action is the most efficient. So the range of the parameter is re-discretized. The 
pseudocode of the proposed method for the case when $i = 2$ is presented in Algorithm~\ref{adaptivelisting}.
\begin{algorithm}[h!]
    \caption{Algorithm with adaptive selection of action set}
    \label{adaptivelisting}
    \begin{algorithmic}[1]
    \STATE {State $s \gets single\_state$}
    \FOR {each parameter $v_i$ to be adjusted}
      \STATE {$P_i \gets \{[v_i^{\min}, v_i^{\max}]\}$}
      \STATE {$A_i \gets $ actions corresponding to partition $P_i$, where $A_i$ is a set of actions for $\text{agent}_i$ 
adjusting parameter $v_i$ }
      \STATE {$Q_i(s, a) \gets 0$}
    \ENDFOR
    \FOR {each parameter $v_i$ to be adjusted}
	\IF {$P_i = \{[v_i^{\min}, v_i^{\max}]\}$}
	  \STATE {\textbf{Split of range ($v_i$)}}
	\ELSIF {$A_i$ contains two or more actions and $Q(s, a) - Q(s, a') < \delta$}
	  \STATE {\textbf{Split of range ($v_i$)}}  
	\ENDIF  
	\STATE {$\text{Agent}_i$ selects action $a_i$ from $A_i$ using $\varepsilon$-greedy strategy} 
    \ENDFOR
    \STATE {Apply actions $a_1 \ldots a_n$ to environment, obtain reward $r$}
    \FOR {each selected action $a_i$}
	\STATE {$Q(s, a_i) \gets Q(s,a_i)+\alpha(r+\gamma\max\limits_{a_i'}Q(s,a_i')-Q(s,a_i))$}
    \ENDFOR

    \end{algorithmic}
    
    \textbf{Split of range of $v_i$}
    \begin{algorithmic}[1]
    \STATE {Sort saved tuples of $(v, r)$ according to $v_i$}
    \STATE {$tuples\_count \gets$ number of tuples}
    \STATE {$best  \gets +\infty$}
    \FOR {$j \gets 1$ to $tuples\_count$}
	    \STATE {$v_{i,j} \gets$ value of $v_i$ in j-th tuple in sorted list}
	    \STATE {$candidate \gets \frac{v_{i,j} + v_{i, j + 1}}{2}$}
	    \FOR {$j \gets 1$ to $tuples\_count$}
		\STATE {$x \gets \{ r | v_{i, j} \leq candidate\}$}
		\STATE {$y \gets \{ r | v_{i, j} > candidate\}$}
	    \ENDFOR
	    \STATE {Calculate $p$-value for $x$ and $y$ using Kolmogorov-Smirnov criterion}
	    \IF {$p$-value$ < best$}
		    \STATE $best \gets p$-value
		    \STATE $best\_split \gets candidate$
	    \ENDIF
    \ENDFOR	
    \IF {$best > 0.05$}
	\STATE {Split saved tuples into sets $L$ and $R$ according to $best\_split$}
	\STATE {Find split point $s_l$ for $L$}
	\STATE {Find split point $s_r$ for $R$}
	\IF {Split points $s_l$ and $s_r$ are not found}
	    \STATE {$P_i \gets \{[v_{\min}, s], (s, v_{\max}]\}$}
	\ELSIF {split point $s_l$ is not found}
	    \STATE {$P_i \gets \{[v_{\min}, s], (s, s_r], (s_r, v_{\max}]\}$}
	\ELSIF {split point $s_r$ is not found}
	    \STATE {$P_i \gets \{[v_{\min}, s_l], (s_l, s], (s, v_{\max}]\}$}
	\ELSE
	    \STATE {$P_i \gets \{[v_{\min}, s_l], (s_l, s], (s, s_r], (s_r, v_{\max}]\}$}
	\ENDIF
	\STATE {Set $A_i \gets$ actions corresponding to partition $P_i$}
    \ENDIF 
    \end{algorithmic}
\end{algorithm}

\section{Experiments and results}
The proposed methods were compared with the EARPC algorithm, the approach proposed by Karafotias et al. and the $Q$-learning algorithm. In the 
$Q$-learning algorithm a single state is used and the ranges of parameter values are discretized a priori on five 
equally sized intervals as in the 
algorithm proposed by Karafotias et al. The considered methods were tested on several real-valued functions with 
different landscapes and different number of local optima. We 
implemented the EARPC algorithm ourselves and we 
used the implementation of the method proposed by Karafotias et al. kindly given by the authors of~\cite{karafotias-gecco}.
\subsection{Experiment description}
Let us denote the optimized function as $F(x_1, \ldots, x_n) : R^n \rightarrow R$, $x_i \in [min_i, max_i]$. Then the goal of EA is to find a 
vector $x_1 \ldots x_n$, such as the global minimum of the function with $\epsilon$ precision is reached on this vector. The algorithms were tested on the 
sphere function (Eq.~\ref{sphere_eq}), the Rosenbrock function (Eq.~\ref{rosenbrock_eq}), the Levi function (Eq.~\ref{levi_eq}) and the Rastrigin function 
(Eq.~\ref{rastrigin_eq}). 
\begin{equation}
\label{sphere_eq}
f(x_1,..,x_n) = \sum\limits_{i=1}^n{x_i^2}
\end{equation} 
\begin{equation}
\label{rosenbrock_eq}
f(x_1, x_2) = (a - x_1^2)^2 + b(x_2 - x_1^2)^2
\end{equation}
\begin{equation}
\label{levi_eq}
\begin{split}
f(x_1, x2) = &\sin^2(3\pi x) + (x - 1)^2(1 + \sin^2(3\pi y)) + \\
& + (y - 1)^2(1 + \sin^2(2\pi y))
\end{split}
\end{equation}
\begin{equation}
\label{rastrigin_eq}
f(x_1, .., x_n) = A \cdot n + \sum\limits_{i = 1}^n\left[ x_i^2 - A\cos\left(2 \pi x_i \right)\right]
\end{equation}

An individual of EA is a real vector $x_1 \ldots x_n$.  Mutation operator is defined as in the work by Karafotias et al. For each $x_i$ in a vector 
we apply the following transformation:
\begin{equation}
\label{mutationtrans}
x_i = \begin{cases}
	max_i, \text{if } x_i + \sigma dx_i > max_i \\
	min_i, \text{if } x_i + \sigma dx_i < min_i \\
	x_i + \sigma dx_i, \text{in other cases}
      \end{cases}
\end{equation}
where $1 \le i \le n$, $dx_i \sim \mathcal{N}(0, 1)$, and $\sigma$ is the parameter to be adjusted.
We expect that the closer to the global optimum EA is, the smaller $\sigma$ should be. The range of $\sigma$ is $[0, k]$, where $k$ is a constant. As the range 
of $\sigma$ grows, it becomes harder to find the optimal value of $\sigma$.We used different values of $k$ presented in Table~\ref{eaparameters}. Note that 
the higher $k$ is, the greater range of $\sigma$ is.

We used $(\mu + \lambda)$ evolution strategy with different values of $\mu$ and $\lambda$ presented in Table~\ref{eaparameters}. All the considered 
algorithms were run 30 times on each problem instance, then the results were averaged. Parameters of EA are presented in Table~\ref{eaparameters}. 

The reward function in reinforcement learning was defined as follows: $r = c \cdot (\frac{f_t-f_{t + 1}}{f_{t + 1}})$, where $f_t$ is the minimal fitness 
function value in the generation $t$ and $c$ is a constant taken from~\cite{karafotias-gecco}. Value of $c$ is presented in Table~\ref{learningparameters}. 
Note that when we solve minimization problem with elitist selection, $f_{t + 1}$ is less than or equal to $f_t$, so the reward is always positive. Parameters 
of reinforcement learning were taken from~\cite{karafotias-gecco}. They are presented in Table~\ref{learningparameters}.
\begin{table}[ht]
\begin{center}
\caption{EA parameters} \label{eaparameters}
\begin{tabular}{|lll|}
\hline
Parameter & Description & Value\\
\hline
$k$ & maximal value of $\sigma$ & $\{1, 2, 3\}$\\
$\mu$ & the number of parents & $\{1, 5, 10\}$\\
$\lambda$ & the number of children & $\{1, 3, 7\}$ \\
$\epsilon$ & precision & $10^{-5}$\\ \hline
\end{tabular}
\end{center}
\end{table}

\begin{table}[ht]
\begin{center}
\caption{RL parameters} \label{learningparameters}
\begin{tabular}{|lll|}
\hline
Parameter & Description & Value\\
\hline
$\alpha$ & learning rate & 0.9\\
$\alpha_0$ & learning rate in case of zero reward & 0.02 \\
$\gamma$ & discount factor & 0.8\\
$\varepsilon$ & exploration probability & 0.1\\
$c$ & coefficient in reward & 100 \\ \hline
\end{tabular}
\end{center}
\end{table}



\subsection{Results and discussion}

\begin{table*}[t]
  \caption{Averaged number of runs needed to reach the optimum using the the proposed method with adaptive selection of action set (A), the Q-learning 
algorithm (Q), the approach proposed by Karafotias et al. (K), the EARPC algorithm (E) and the proposed method which combines Karafotias et al. and EARPC 
(E+K)}    
\small
\begin{tabular}{|p{0.15em}|p{0.3em}|p{0.15em}|p{1.1em}|p{1.5em}|p{1.5em}|p{1.5em}|p{1.6em}|p{1.1em}|p{1.5em}|p{1.5em}|p{1.5em}|p{1.6em}|p{1.1em}|p{1.5em}|p{
1.5em } |p { 1.5em } |p{ 1.6em} |p{1.1em}|p{1.5em}|p{1.5em}|p{1.5em}|p{1.6em}|}
    \hline
    & & & \multicolumn{5}{c|}{Sphere function} & \multicolumn{5}{c|}{Rastrigin function} & \multicolumn{5}{c|}{Levi function} & 
\multicolumn{5}{c|}{Rosenbrock function}\\
    \hline
     \ind$k$&\ind$\mu$&\ind$\lambda$ & A                            & Q            & K                          & E          & E+K                 
                                     & A                            & Q            & K                          & E          & E+K
                                     & A                            & Q            & K                          & E          & E+K 
                                     & A                            & Q            & K                          & E      
    & E+K\\ \hline
    \ind 1   &   \ind 1	 & \ind 1    & \ind \cellcolor{res}{2434} & \ind 8769    & \ind 8048                  & \ind 5258  & \ind 4830                   
                                     & \ind \cellcolor{res}{3631} & \ind 9653    & \ind 8385 		        & \ind 7794  & \ind 8689  
                                     & \ind \cellcolor{res}{3496} & \ind 7200    & \ind 7265                  & \ind 7986  & \ind 14092 
                                     & \ind \cellcolor{res}{5124} & \ind 15058   & \ind 13418                 & \ind 9003  & \ind 12905 \\
    \ind 1   &   \ind 1	 & \ind 3    & \ind \cellcolor{res}{2207} & \ind 4221    & \ind 2683                  & \ind 3942  & \ind 3070                   
                                     & \ind \cellcolor{res}{1776} & \ind 2226    & \ind 2069                  & \ind 3148  & \ind 3610 
                                     & \ind \cellcolor{res}{1980} & \ind 3305    & \ind 2903                  & \ind 2789  & \ind 3688 
                                     & \ind \cellcolor{res}{2301} & \ind 3914    & \ind 4167                  & \ind 3553  & \ind 2701 \\ 
    \ind 1   &   \ind 1	 & \ind 7    & \ind \cellcolor{res}{878}  & \ind 1085    & \ind 2620                  & \ind 1653  & \ind 2247                   
                                     & \ind \cellcolor{res}{1226} & \ind 1605    & \ind 1757                  & \ind 1422  & \ind 1422 
                                     & \ind \cellcolor{res}{1321} & \ind 1584    & \ind 1600                  & \ind 1923  & \ind 3820
                                     & \ind \cellcolor{res}{1411} & \ind 1791    & \ind 2330                  & \ind 2296  & \ind 1941\\
    \ind 1   &   \ind 5	 & \ind 1    & \ind \cellcolor{res}{1450} & \ind 1664    & \ind 2076                  & \ind 4472  & \ind 3893                   
                                     & \ind \cellcolor{res}{1666} & \ind 1706    & \ind 2281                  & \ind 4341  & \ind 4530 
                                     & \ind \cellcolor{res}{1778} & \ind 1898    & \ind 1865                  & \ind 2372  & \ind 2246 
                                     & \ind \cellcolor{res}{1859} & \ind 2311    & \ind 2809                  & \ind 5730  & \ind 4453 \\
    \ind 1   &   \ind 5	 & \ind 3    & \ind \cellcolor{res}{569}  & \ind 706     & \ind 824                   & \ind 1589  & \ind 845                    
                                     & \ind 918                     & \ind \cellcolor{res}{894} & \ind 942    & \ind 1958  & \ind 2197 
                                     & \ind 855                     & \ind 862     & \ind \cellcolor{res}{794}& \ind 1632  & \ind 2171 
                                     & \ind 1053                    & \ind 869     & \ind \cellcolor{res}{748}& \ind 1393  & \ind 2749 \\
    \ind 1   &   \ind 5	 & \ind 7    & \ind \cellcolor{res}{368}  & \ind 390     & \ind 473                   & \ind 642   & \ind 568                    
                                     & \ind \cellcolor{res}{502}  & \ind 570     & \ind 675                   & \ind 1679  & \ind 1329 
                                     & \ind \cellcolor{res}{356}  & \ind 606     & \ind 444                   & \ind 1426  & \ind 1236 
                                     & \ind \cellcolor{res}{401}  & \ind 533     & \ind 665                   & \ind 1130  & \ind 1258 \\
    \ind 1   &   \ind 10 & \ind 1    & \ind \cellcolor{res}{703}  & \ind 959     & \ind 747                   & \ind 1438  & \ind 728                    
                                     & \ind \cellcolor{res}{1008} & \ind 1103    & \ind 1105                  & \ind 2681  & \ind 2926 
                                     & \ind 884                     & \ind 840     & \ind \cellcolor{res}{804}& \ind 2353  & \ind 1451 
                                     & \ind 1617                    & \ind \cellcolor{res}{1497} & \ind 1593  & \ind 2213  & \ind 3085 \\
    \ind 1   &   \ind 10 & \ind 3    & \ind \cellcolor{res}{331}  & \ind 378     & \ind 358                   & \ind 534   & \ind 400                    
                                     & \ind 622                     & \ind \cellcolor{res}{604} & \ind 665    & \ind 1460  & \ind 1485 
                                     & \ind 533                     & \ind \cellcolor{res}{502} & \ind 624    & \ind 1491  & \ind 1134 
                                     & \ind 683                     & \ind \cellcolor{res}{488} & \ind 616    & \ind 1161  & \ind 1225 \\
    \ind 1   &   \ind 10 & \ind 7    & \ind 186 		    & \ind 195     & \ind \cellcolor{res}{167}& \ind 142   & \ind 422                    
                                     & \ind \cellcolor{res}{358}  & \ind 489     & \ind 473                   & \ind 1103  & \ind 1858 
                                     & \ind 308                     & \ind 331     & \ind \cellcolor{res}{294}& \ind 510   & \ind 766 
                                     & \ind 483                     & \ind 290     & \ind                                      \cellcolor{res}{235}& \ind 391   & \ind 429 \\\hline
    \ind 2   &   \ind 1	 & \ind 1    & \ind \cellcolor{res}{4342} & \ind 25192   & \ind 28523                 & \ind 31738 & \ind 16182                  
                                     & \ind \cellcolor{res}{4744} & \ind 23663   & \ind 21043                 & \ind 27865 & \ind 19142 
                                     & \ind \cellcolor{res}{4947} & \ind 13420   & \ind 14789                 & \ind 25721 & \ind 30653 
                                     & \ind \cellcolor{res}{5165} & \ind 23371   & \ind 27239                 & \ind 20005 & \ind 20819 \\
    \ind 2   &   \ind 1	 & \ind 3    & \ind \cellcolor{res}{2333} & \ind 7681    & \ind 6478                  & \ind 5152  & \ind 4233                   
				     & \ind \cellcolor{res}{1839} & \ind 6405    & \ind 6825                  & \ind 9748  & \ind 8997 
				     & \ind \cellcolor{res}{2020} & \ind 6884    & \ind 6601                  & \ind 7477  & \ind 4612 
				     & \ind \cellcolor{res}{3461} & \ind 7295    & \ind 7169                  & \ind 7166  & \ind 13342 \\
    \ind 2   &   \ind 1	 & \ind 7    & \ind \cellcolor{res}{1464} & \ind 3360    & \ind 3739                  & \ind 1688  & \ind 2956                   
				     & \ind \cellcolor{res}{1160} & \ind 2388    & \ind 2183                  & \ind 3806  & \ind 4085
			             & \ind \cellcolor{res}{1205} & \ind 3521    & \ind 2370                  & \ind 2533  & \ind 2967 
			             & \ind \cellcolor{res}{1753} & \ind 3488    & \ind 2968                  & \ind 5874  & \ind 7870\\
    \ind 2   &   \ind 5	 & \ind 1    & \ind \cellcolor{res}{1891} & \ind 3814    & \ind 3468                  & \ind 4908  & \ind 5173                   
				     & \ind \cellcolor{res}{2467} & \ind 3944    & \ind 4029                  & \ind 5961  & \ind 5750 
				     & \ind \cellcolor{res}{1935} & \ind 3517    & \ind 3369                  & \ind 7222  & \ind 5365 
				     & \ind \cellcolor{res}{1990} & \ind 8076    & \ind 5810                  & \ind 11153 & \ind 11856 \\
    \ind 2   &   \ind 5	 & \ind 3    & \ind 974                     & \ind 994     & \ind 1163                  & \ind 1631  & \ind \cellcolor{res}{946} 
			             & \ind \cellcolor{res}{1128} & \ind 1614    & \ind 1780                  & \ind 2293  & \ind 1720 
				     & \ind \cellcolor{res}{1085} & \ind 1231    & \ind 1326                  & \ind 3039  & \ind 2943 
				     & \ind \cellcolor{res}{1396} & \ind 2116    & \ind 2705                  & \ind 3775  & \ind  4542 \\
    \ind 2   &   \ind 5	 & \ind 7    & \ind \cellcolor{res}{616}  & \ind 716     & \ind 756                   & \ind 1511  & \ind 626                    
				     & \ind 967                     & \ind 1012    & \ind 1461                  & \ind \cellcolor{res}{933} & \ind 1180
				     & \ind \cellcolor{res}{770}  & \ind 792     & \ind 1089                  & \ind 1153  & \ind 2180 
				     & \ind \cellcolor{res}{934}  & \ind 1001    & \ind 1247                  & \ind 1775  & \ind 2584\\
    \ind 2   &   \ind 10 & \ind 1    & \ind 1164 		    & \ind 2397    & \ind 1833                  & \ind \cellcolor{res}{778} & \ind 876   
				     & \ind \cellcolor{res}{1722} & \ind 2108    & \ind 2255                  & \ind 2411  & \ind 2807 
				     & \ind \cellcolor{res}{1803} & \ind 1884    & \ind 2197                  & \ind 5139  & \ind 3072 
				     & \ind \cellcolor{res}{2022} & \ind 3415    & \ind 2787                  & \ind 8768  & \ind 4603 \\
    \ind 2   &   \ind 10 & \ind 3    & \ind 459		            & \ind \cellcolor{res}{445} & \ind 465    & \ind 719   & \ind 788                    
				     & \ind \cellcolor{res}{988}  & \ind 1420    & \ind 1116                  & \ind 1486  & \ind 1234 
				     & \ind \cellcolor{res}{887}  & \ind 988     & \ind 1006                  & \ind 1045  & \ind 2381 
				     & \ind 1181                    & \ind \cellcolor{res}{1037} & \ind 1106  & \ind 3719  & \ind 1648 \\
    \ind 2   &   \ind 10 & \ind 7    & \ind \cellcolor{res}{188}  & \ind 320     & \ind 252                   & \ind 380   & \ind 413                    
				     & \ind \cellcolor{res}{615}  & \ind 856     & \ind 712                   & \ind 627   & \ind 922  
				     & \ind 618                     & \ind 731     & \ind \cellcolor{res}{564}& \ind 806   & \ind 1064 
				     & \ind 707                     & \ind 811     & \ind \cellcolor{res}{666}& \ind 1740  & 				     \ind 2558\\\hline
    \ind 3   &   \ind 1  & \ind 1    & \ind \cellcolor{res}{8055} & \ind 36698   & \ind 29112                 & \ind 54868 & \ind 30710                  
				     & \ind \cellcolor{res}{5159} & \ind 29082   & \ind 23305                 & \ind 24249 & \ind 27327 
				     & \ind \cellcolor{res}{5216} & \ind 21470   & \ind 21953                 & \ind 28900 & \ind 35119 
				     & \ind \cellcolor{res}{6535} & \ind 28702   & \ind 36597                 & \ind 32533 & \ind 43832 \\
    \ind 3   &   \ind 1	 & \ind 3    & \ind \cellcolor{res}{2826} & \ind 7845    & \ind 9115                  & \ind 12446 & \ind 11985                  
				     & \ind \cellcolor{res}{2821} & \ind 7977    & \ind 10726                 & \ind 6541  & \ind 16040 
				     & \ind \cellcolor{res}{2900} & \ind 9029    & \ind 7071                  & \ind 6966  & \ind 10883 
				     & \ind \cellcolor{res}{3391} & \ind 11427   & \ind 9873                  & \ind 13061 & \ind 10068\\
    \ind 3   &   \ind 1	 & \ind 7    & \ind \cellcolor{res}{1427} & \ind 3907    & \ind 4813                  & \ind 10118 & \ind 9207                   
				     & \ind \cellcolor{res}{1517} & \ind 4680    & \ind 4266                  & \ind 4144  & \ind 5408 
				     & \ind \cellcolor{res}{1700} & \ind 4139    & \ind 4019                  & \ind 3375  & \ind 2062  
				     & \ind \cellcolor{res}{2573} & \ind 5820    & \ind 6462                  & \ind 9797  & \ind 9775\\
    \ind 3   &   \ind 5	 & \ind 1    & \ind \cellcolor{res}{2447} & \ind 8328    & \ind 5886                  & \ind 3348  & \ind 12313                  
				     & \ind \cellcolor{res}{2704} & \ind 6208    & \ind 5419                  & \ind 5953  & \ind 7665 
				     & \ind \cellcolor{res}{3105} & \ind 6966    & \ind 6742                  & \ind 10480 & \ind 9059 
				     & \ind \cellcolor{res}{3148} & \ind 12438   & \ind 6791                  & \ind 8952  & \ind 11581 \\
    \ind 3   &   \ind 5	 & \ind 3    & \ind 1445                    & \ind 2790    & \ind 2222                  & \ind \cellcolor{res}{1379} & \ind 1848 
				     & \ind \cellcolor{res}{1544} & \ind 2514    & \ind 2096                  & \ind 2823  & \ind 2304 
			             & \ind \cellcolor{res}{1808} & \ind 2467    & \ind 2664                  & \ind 4593  & \ind 3714 
			             & \ind \cellcolor{res}{1721} & \ind 2289    & \ind 3673                  & \ind 4434  & \ind 6799 \\
    \ind 3   &   \ind 5	 & \ind 7    & \ind 896                     & \ind 919     & \ind \cellcolor{res}{777}& \ind 1424  & \ind 1105                   
				     & \ind \cellcolor{res}{902}  & \ind 1120    & \ind 1629                  & \ind 1033  & \ind 1055
				     & \ind 1017                    & \ind 1929    & \ind 1788                  & \ind \cellcolor{res}{594} & \ind 1439
				     & \ind \cellcolor{res}{1231} & \ind 1626    & \ind 1255                  & \ind 3951  & \ind 4914\\
    \ind 3   &   \ind 10 & \ind 1    & \ind \cellcolor{res}{1996} & \ind 2398    & \ind 2531                  & \ind 3173  & \ind 3745                   
				     & \ind \cellcolor{res}{2048} & \ind 3747    & \ind 3258                  & \ind 5095  & \ind 3873
				     & \ind \cellcolor{res}{2071} & \ind 2901    & \ind 2943                  & \ind 5210  & \ind 4814 
				     & \ind \cellcolor{res}{2352} & \ind 4473    & \ind 6531                  & \ind 8320  & \ind 10539 \\
    \ind 3   &   \ind 10 & \ind 3    & \ind \cellcolor{res}{1053} & \ind 1074    & \ind 1206                  & \ind 1114  & \ind 171                    
				     & \ind 1296                    & \ind 1740    & \ind 1523                  & \ind \cellcolor{res}{1013} & \ind 1028
				     & \ind \cellcolor{res}{1330} & \ind 1462    & \ind 1832                  & \ind 3140  & \ind 2389 
				     & \ind 1677                    & \ind 2000    & \ind \cellcolor{res}{1650}& \ind 2611 & \ind 2479 \\
    \ind 3   &   \ind 10 & \ind 7    & \ind 409                     & \ind \cellcolor{res}{391} & \ind 427    & \ind 410   & \ind 537                    
				     & \ind 955                     & \ind 995     & \ind 1203                  & \ind \cellcolor{res}{779} & \ind 839
				     & \ind 667                     & \ind 1117    & \ind 799                   & \ind \cellcolor{res}{521} & \ind 845
				     & \ind 1101                    & \ind 1137    & \ind \cellcolor{res}{1098}& \ind 
1742 & \ind 1907\\ \hline
   \multicolumn{3}{|p{0.3em}|}{\ind Summary}& \cellcolor{res}20			    & 2            & 2           
               & 2          & 1
				     & \cellcolor{res}22			    & 2            & 0                   
       & 3          & 0
				     & \cellcolor{res}19                           & 1            & 4                  
        & 2          & 0
				     & \cellcolor{res}19                           & 3            & 5                  
        & 0          & 0 \\
    \hline
  \end{tabular}
  \label{results}
\end{table*}

The results are presented as follows. First, we present and analyze the obtained average number of generations needed to reach the optimum using the two 
proposed methods, the EARPC algorithm, the approach proposed by Karafotias et al. and the $Q$-learning algorithm. Second, we analyze the selected values 
of the adjusted parameter. Also for the algorithm with dynamic discretization of the parameter range we analyze values of selected split points.

\subsubsection{Influence on efficiency of EA}
The average number of generations needed to reach the optimum using different parameter controllers is presented in Table~\ref{results}. The first three 
columns contain values of EA parameters $k$, $\mu$ and $\lambda$ correspondingly. The next 20 columns contain results of optimizing the following functions: 
the sphere function, the Rastrigin function, the Levi function and the Rosenbrock function. In each run we adjust $\sigma$, the parameter of mutation from 
Eq.~\ref{mutationtrans}. We 
expect that the closer to the global optimum EA is, the smaller $\sigma$ should be. For each function, we present the results of the following parameter 
controllers: the proposed method with adaptive selection of action set (A), the Q-learning algorithm (Q), the approach 
proposed by Karafotias et al. (K), the EARPC algorithm (E) and the proposed method which combines Karafotias et al. and EARPC (E+K). In the 
last row, for each algorithm the total number of the EA configurations on which this algorithm outperformed all other algorithms is presented. 

For each problem instance, the average number of generations needed to reach the optimal value is presented. The gray background corresponds to the best result 
for each EA configuration. Note that we do not compare EA algorithms characterized by different values of $\mu$ and 
$\lambda$. We only compare methods of parameter control on the same EA configuration. For each set of $k$, $\mu$, $\lambda$ values the number of fitness 
function evaluations in a generation is the same for all parameter control methods. So using the number of generations instead of the number of fitness 
evaluations does not affect results of comparison within a row.

The deviation of the average number of generations needed to reach the optimum in the proposed method with adaptive selection of action set is less than 
5\%. For the Q-learning algorithm and the approach proposed by Karafotias et al. the deviation is about 10\%. For the EARPC algorithm and 
the proposed method which combines Karafotias et al. and EARPC the deviation is about 40\%.

\begin{figure*}[t]
  \centering
  \subfloat[]{\includegraphics[width=0.32\textwidth]{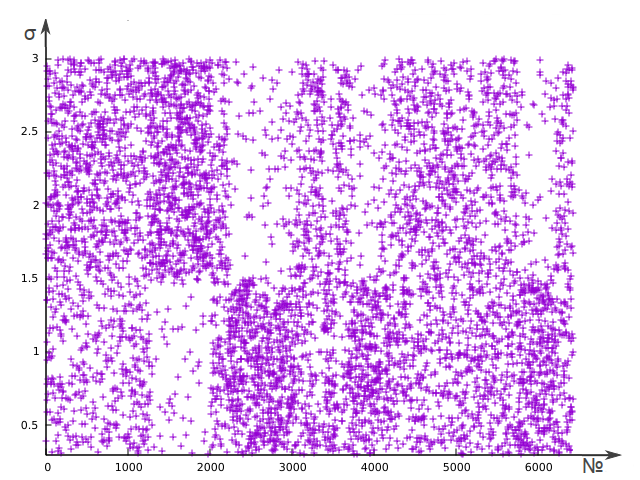} \label{erd}}
  \subfloat[]{\includegraphics[width=0.32\textwidth]{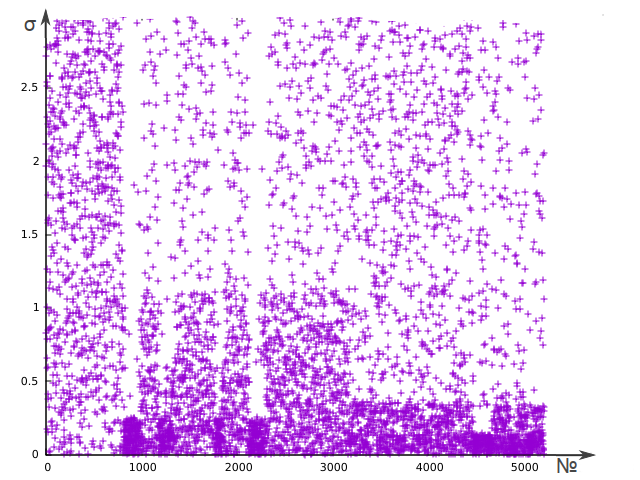} \label{ard}}
  \subfloat[]{\includegraphics[width=0.32\textwidth]{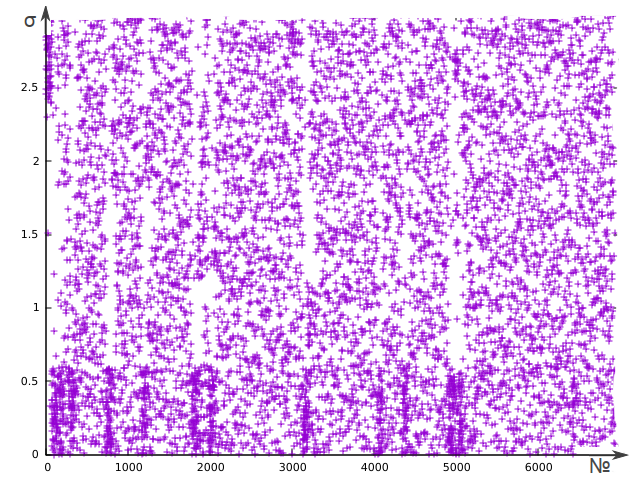} \label{kr}}
   \caption{Selected values of $\sigma$ in the proposed method which combines Karafotias et al. and EARPC (a), the proposed method with 
adaptive selection of action set (b) and the method proposed by Karafotias et al. (c) on Rastrigin function.}
  \label{selectedvals}
\end{figure*}

\begin{figure*}[t]
  \centering
  \subfloat[]{\includegraphics[width=0.49\textwidth]{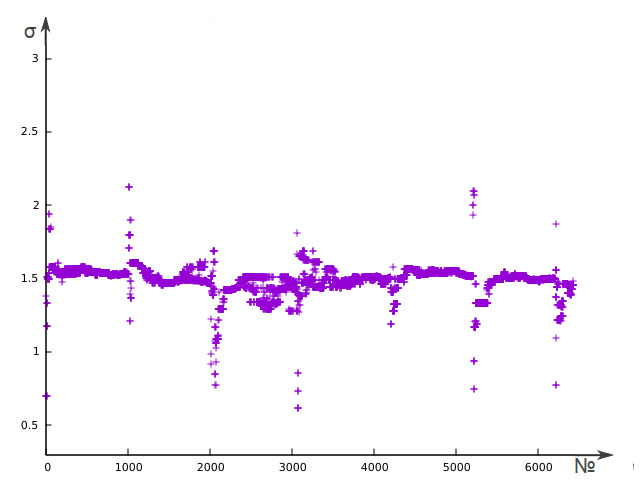} \label{er}}
  \subfloat[]{\includegraphics[width=0.49\textwidth]{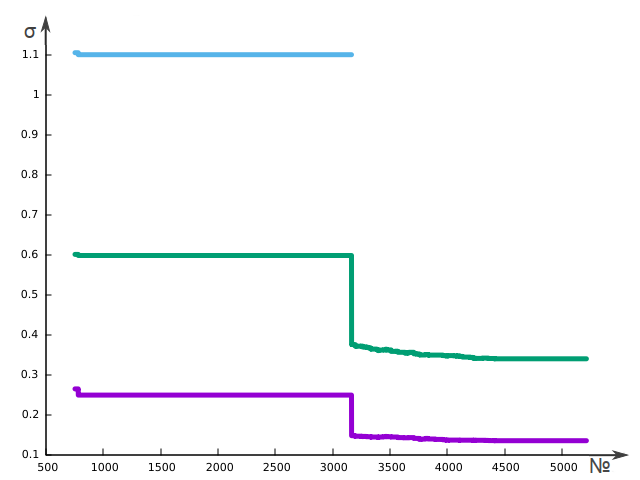} \label{ar}}
  \caption{Split point values in the proposed method which combines Karafotias et al. and EARPC (a) and the proposed method with adaptive 
selection of action set (b) on Rastrigin function.}
  \label{selectedpoints}
\end{figure*}


Overall, when $k$ increases, the range of $\sigma$ increases and selection of the optimal value of $\sigma$ becomes 
harder.  
The proposed method with adaptive selection of action set outperformed all other considered methods on most problem instances. According to multiple sign 
test~\cite{stattest}, the proposed method with adaptive selection of action set is distinguishable from the other methods at the level of statistical 
significance 
$\alpha = 0.05$.

Let us discuss the possible reasons why the proposed method with adaptive selection of action set outperformed the other considered methods. 
In the method proposed by Karafotias et al., in the case of zero reward, for some consecuent iterations $Q$-values become zero. So the 
experience of the agent is lost. The proposed method with adaptive selection of action set does not have this 
drawback because experience is not saved only in $Q$-values. 

In EARPC and the proposed method which combines Karafotias et al. and EARPC, the range of 
parameter $\sigma$ is split into two almost equal intervals. The proposed method with adaptive 
selection of action set splits interval using another criterion which allows to split the range more 
precisely.
  
Consider the efficiency of using binary decision tree of states. The $Q$-learning algorithm and the algorithm proposed by Karafotias et al. demonstrate 
similar performance. The same is also true for the EARPC algorithm and the proposed method which combines Karafotias et al. and EARPC. Therefore, applying 
dynamic 
state space segmentation algorithms in parameter controllers does not seem to increase efficiency of EA when solving the considered problems.


\newpage
\subsubsection{Selected parameter values and split points}

In the Fig.~\ref{selectedvals} values of $\sigma$ selected by the two proposed methods and the method proposed by Karafotias et al. during Rastrigin 
function optimization are presented.
For the other considered functions plots are similar. For brevity, they are not presented. The horizontal axis refers to the number of an iteration, the 
vertical 
axis refers to the selected value of $\sigma$. 

We can see that method which combines Karafotias et al. and EARPC~(Fig.~\ref{erd})  in the end of 
optimization 
selects $\sigma$ almost randomly.
The algorithm proposed by Karafotias et al.~(Fig.~\ref{kr}) continues selection of an action if it has 
achieved positive reward for application of 
this 
action. However, if the algorithm has not obtained positive reward in some consequent iterations, it loses information 
about efficient action selected earlier. 
 We can see that the proposed method with adaptive selection of action set~(Fig.~\ref{ard}) in the beginning 
of the optimization 
selects $\sigma$ randomly, but during the optimization process the selected $\sigma$ convergences to the optimal value.

For the two proposed methods, the selected split points are shown in Fig.~\ref{selectedpoints}. The algorithm proposed by Karafotias et al. discretizes the 
parameter range a priori, so the considered type of plot is not applicable. The split points selected by EARPC are not presented because EARPC and the first 
proposed method demonstrate similar performance.
The horizontal axis in Fig.~\ref{selectedpoints} refers to the 
number of an iteration, the vertical axis refers to the selected split points of the $\sigma$ range. 

We can see that in the proposed method which combines Karafotias et 
al. and EARPC~(Fig.~\ref{er}), the parameter range is split into two almost equal intervals on each 
iteration of EA. The proposed method with 
adaptive selection of action set~(Fig.~\ref{ar}) does not split interval until enough tuples of experience are 
obtained. Then (after about 
750 iterations) the $\sigma$ range is split into four intervals by three split points. Discretization of the 
$\sigma$ range is not changed until 
$Q$-values become almost equal for all actions (after about 3150 iterations). Then the range is discretized again into
three intervals. Note that the 
intervals after re-discretization are shrunk and the number of intervals is reduced. So the agent can select a good 
value of $\sigma$ more precisely. This 
effect is reflected in Fig.~\ref{ard}. After about 3150 iterations of the algorithm the agent almost always 
selects the value of $\sigma$ close to 
the optimal value. The outliers can be explained by the fact that $\varepsilon$-greedy exploration strategy is used in $Q$-learning. According to this strategy, 
a random action is applied with probability $\varepsilon$.


\newpage
\section{Conclusion}

We proposed two new parameter controllers based on reinforcement learning. These algorithms discretize parameter range 
dynamically. One of the proposed methods 
is based on two existing parameter controllers: EARPC and the algorithm proposed by Karafotias et al. In the second 
approach the parameter range is discretized using Kolmogorov-Smirnov criterion and it is re-discretized if the 
expected reward is almost equal for all actions of the agent.   

The proposed methods were compared with EARPC, the algorithm proposed by Karafotias et al. and the Q-learning algorithm. We 
tested controllers with 27 configurations of  EA on four test problems. On the most problem instances, the second proposed approach outperformed the other 
considered methods. We showed that this method improves the parameter value during the whole optimization process contrary to the other methods. 
It also may be noticed that application of dynamic state space segmentation algorithms in parameter controllers does not seem to increase efficiency of EA when 
solving the considered test problems.      

 \bibliographystyle{abbrv}
 \bibliography{../../../bibliography}
\end{document}